\documentclass[10pt,twocolumn,letterpaper]{article}

\usepackage{cvpr}
\usepackage{times}
\usepackage{epsfig}
\usepackage{graphicx}
\usepackage{amsmath}
\usepackage{amssymb}
\usepackage{subfigure}
\usepackage{multirow}

\usepackage[breaklinks=true,bookmarks=false]{hyperref}

\cvprfinalcopy 

\def\cvprPaperID{****} 

\begin{document}

\title{vireoJD-MM at Activity Detection in Extended Videos}

\author{Fuchen Long, Qi Cai, Zhaofan Qiu, Zhijian Hou, Yingwei Pan, Ting Yao and Chong-Wah Ngo\\
         City University of Hong Kong, Kowloon, Hong Kong\\JD AI Reseach, Beijing, China\\
         {\tt\small cwngo@cs.cityu.edu.hk, panyw.ustc@gmail.com}
}

\maketitle

\begin{abstract}
This notebook paper presents an overview and comparative analysis of our system designed for activity detection in extended videos (ActEV-PC) in ActivityNet Challenge 2019. Specifically, we exploit person/vehicle detections in spatial level and action localization in temporal level for action detection in surveillance videos. The mechanism of different tubelet generation and model decomposition methods are studied as well. The detection results are finally predicted by late fusing the results from each component.

\section{Introduction}
Action/activity recognition~\cite{tran2014learning,qiu2017deep,qiu2019learning,qiu2019trimmed} and localization~\cite{shou2016temporal,Shou:CVPR17,Lin:ARXIV17,long2019gaussian,awad2017video,zhang2016vireo} in videos is a challenging task for video understanding as video is an information-intensive media with complex variations. The challenge mainly originates from the sparsity of actions at both spatial and temporal dimensions, especially for temporal action detection in surveillance video. In particular, almost half of the videos contain 30 second of footage where no actions are performed. Moreover, the bounding boxes of actions are very small. For example, when the action of ``Riding" is occurring, it only takes up on average less than 2.6\% of the pixels in each frame. Moreover, the similarity of each action and each environment makes it very difficult to use the context of the surrounding scene to assist in classification.

In this work, we aim at localizing action tubelet in spatial level and detecting the action segment in temporal level. Moreover, the tubelet generation methods like greedy linking or tracking are also investigated for boosting the following action recognition and localization. The model decomposition to localize vehicle and person actions are further explored to enhance final performance.

The remaining sections are organized as follows.  We describe the whole framework of our system in section 2. Section 3 presents the object detection model to detect the associated objects with regard to the contained activities and Section 4 details the tubelet generation methods for spatial action localization. Then the descriptions of activities localization and empirical evaluations of whole system are provided in Section 5, followed by conclusion in Section 6.

\begin{table}[!tb]
\centering
\caption{Car-related and Person-related action categories.}
\vspace{0.05in}
\label{table:1.1}
\begin{tabular}{|c|c|}\hline
\textbf{Car-related category}&\textbf{Person-related category} \\ \hline
\text{Closing}              &\text{specialized\_talking\_phone}   \\ \hline
\text{Opening}              &\text{specialized\_texting\_phone}   \\ \hline
\text{Closing\_Trunk}        &\text{Transport\_HeavyCarry}   \\ \hline
\text{Open\_Trunk}           &\text{activity\_carrying}   \\ \hline
\text{vehicle\_turning\_left} &\text{Pull}   \\ \hline
\text{vehicle\_turning\_right}&\text{Riding}   \\ \hline
\text{vehicle\_u\_turn}       &\text{Talking}   \\ \hline
\text{Entering}            &\text{Loading}   \\ \hline
\text{Exiting}              &\text{Unloading}   \\ \hline
\end{tabular}
\end{table}

\begin{table*}
\centering
\caption{Comparison of different tubelet generation and model decomposition methods in our system for activities detection in extended videos task.}
\vspace{0.1in}
\label{table:1.2}
\begin{tabular}{|c|c|c|c|} \hline
\multicolumn{4}{|c|}{Validation Set} \\ \hline
\textbf{Tubelet Generation}~~~&~~~\textbf{Backbone}~~~&~~~\textbf{Model Decomposition}~~~&~~~\textbf{mean-w\_p\_miss@0.15rfa}~~~\\ \hline
Greedy Linking  & P3D-B & No model decomposition & 85.11\%\\
Tracking        & P3D-B & No model decomposition & 84.13\%\\
Tracking        & P3D-B & Vehicle-related model  & 75.62\%\\
Tracking        & P3D-B & Person-related model   & 76.77\%\\
Tracking        & P3D-B & Model decomposition (all) & 76.09\%\\
Tracking        & P3D-B & Model decomposition (all) + softNMS & \textbf{75.17}\%\\ \hline
\multicolumn{4}{|c|}{Test Set} \\ \hline
\textbf{Tubelet Generation}~~~&~~~\textbf{Backbone}~~~&~~~\textbf{Model Decomposition}~~~&~~~\textbf{mean-w\_p\_miss@0.15rfa}~~~\\ \hline
Tracking        & P3D-B & Model decomposition (all) + softNMS & 76.82\%\\
\hline
\end{tabular}
\end{table*}
\end{abstract}

\section{Framework}
\begin{figure}[!tb]
  \centering {\includegraphics[width=0.5\textwidth]{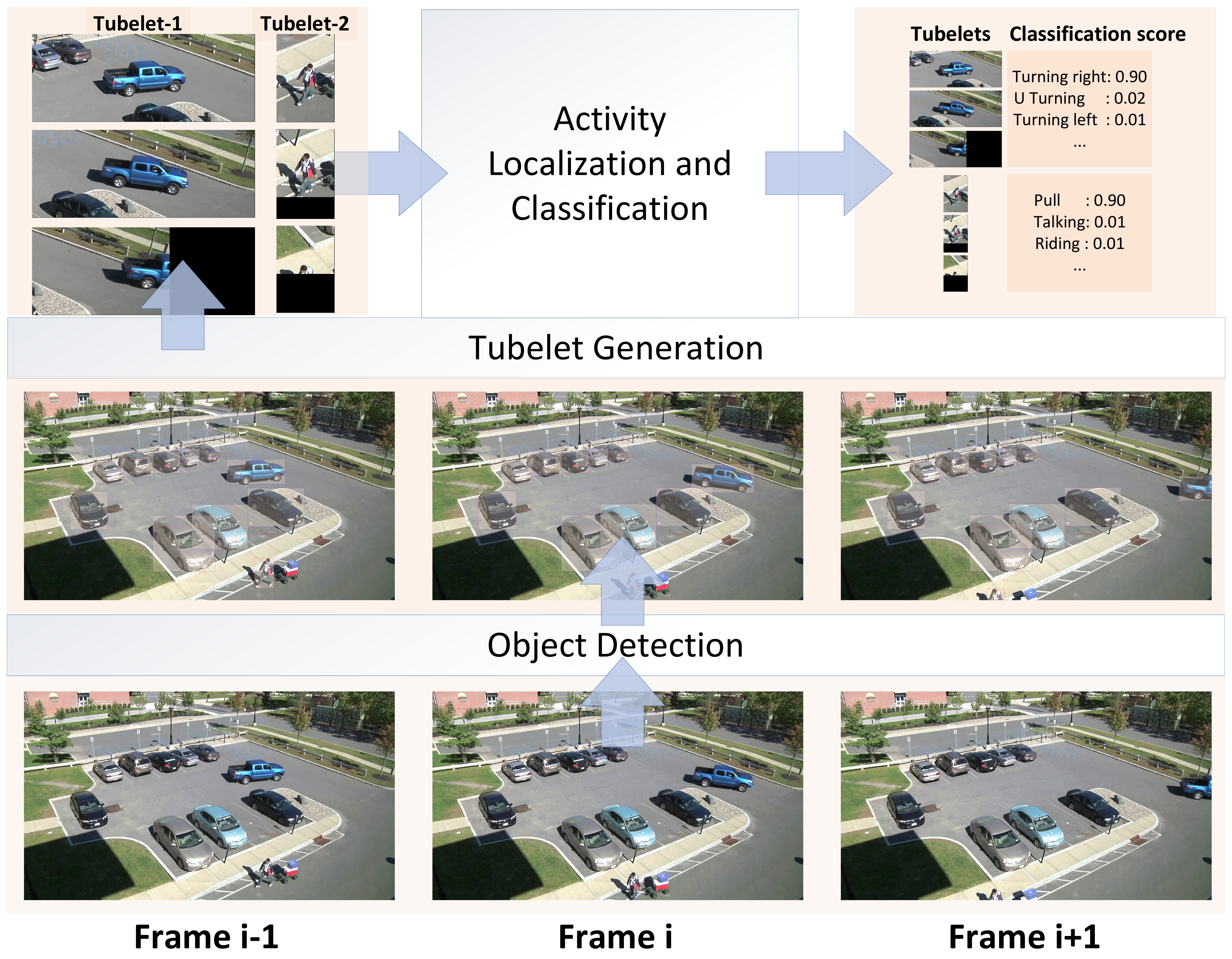}}
  \caption{Framework of our three-stage system for Activity Detection in Extended Videos.}
  \label{fig:framework1}
   \vspace{-0.2in}
  \end{figure}

We present a three-stage system to automatically detect and temporally localize all instances of given activities in the video, which consists of object detection \cite{ren2015faster,cai2019exploring}, tubelet generation and activity localization and classification. We first adopt object detection to localize associated objects related to activities in video, and then utilize tubelet generation to link the detected objects into a long-term tubelet. Finally, we temporally localize all activities in each tubelet and classify them into corresponding categories. Figure \ref{fig:framework1} demonstrates the framework of our proposed system.

\section{Object Detection}
Since each activity only appears in a small spatial area in the video and is associated with specific objects, we utilize object detection to detect  associated objects and further get the spatial localization of the activity. To detect the associated objects from video, we use the Mask R-CNN \cite{he2015deep} model, which is developed for object detection and instance segmentation. The Mask R-CNN contains a box head and mask head which outputs bounding box and segmentation mask respectively. We only keep the bounding box since segmentation mask information is not used. The model is trained on COCO \cite{lin2014microsoft} dataset and achieves 42.1\% mmAP on COCO dataset. We tried finetuning the detection model on ActEV dataset but it degrades the performance since the dataset is not fully annotated and the sparse annotations bring too many false negative during training. Finally, we keep the detection boxes belonging to person, car, truck and bicycle since they are related to the activities in ActEV.

\section{Tublet Generation}
To generate tubelts from detection results, we design two strategies for object linking: greedy based object linking and tracking based object linking. Then we perform tubelet refining, tubelet cropping and tubelet jittering to generate enriched tubelets for localization and classification.

\begin{figure}[!tb]
  \centering {\includegraphics[width=0.5\textwidth]{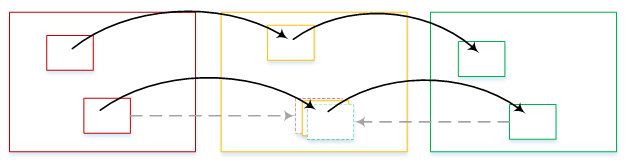}}
  \caption{Greedy based object linking with bounding boxes and scores interpolating.}
  \label{fig:linking1}
   \vspace{-0.2in}
  \end{figure}

\textbf{Greedy based object linking.}
Greedy based object linking starts with interpolating the detection results across frames. Due to that object detection model is not optimal thus there may be many missed detections. For example in Figure \ref{fig:linking1}, the detection model successfully detects two bounding boxes with high confidence in frame $i-1$ and frame $i+1$, while in frame $i$ the yellow box is missed. We interpolate the bounding box coordinates and scores in frame $i -1$ and frame $i +1$ to recover the missed detection in frame $i$. Then we connect boxes in adjacent frames whose spatial IoUs (Intersection over Union) are larger than 0.5 in a greedy manner to form tubelets.

\textbf{Tracking based object linking.}
However, greedy based object linking is not optimal, especially when many objects are missed in detection stage. So we design another tracking based object linking. For each detected object, we perform single object tracking and merge the tracking box with detection box. Technically, when the tracked box matches with any detected box by at least 0.5 IoU, we keep the detected box otherwise the tracked box. Once the detected box has been matched, it is isolated from future matching. The tracking procedure stops when no detection box could be matched for $T$ frames which $T$ is 50 in our experiments.

\textbf{Tubelet refining and cropping.}
Many generated tubelets in linking step are static tubelets without motion, which increases computation cost for further localization and classification as well as increases the false positive risk for activity classification. To filter out static tubelets, we capture the motion information of each tubelet by optical flow and box coordinates. More Specifically, we compute the max value and mean value within each tubelets as well as bounding box coordinates displacements to represent the motion and remove the tubelets with low motion scales.

To ensure that boxes in each tubelet are of the same size, we extend the box in each frame to the maximum box size in the tubelet. Then we enalarge the box size by $T$ times to include more context where $T$ is 1.2 in our experiments.

\textbf{Tubelet jittering.}
Since each tubelet may contain several activities of various durations, we generate more tubelets to cover activities of various durations by tubelet jittering, which slides temporal windows with different size. The windows size is set to \{32, 64, 128, 256\} and temporal stride is 16.

\begin{figure}[!tb]
  \centering {\includegraphics[width=0.5\textwidth]{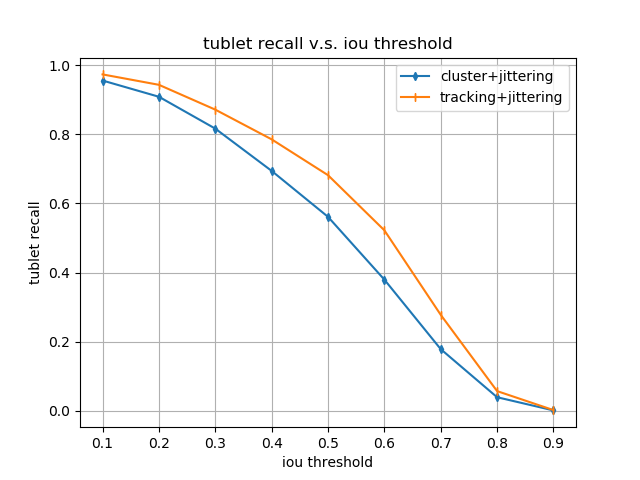}}
  \caption{Recall of tubelet generation with respect to different IoU threshold.}
   \vspace{-0.2in}
  \label{fig:recall}
  \end{figure}

To quantitatively evaluate the performance of tubelet generation, we compute the recall with respect to different IoU threshold. Under the threshold $\tau$, a tubelet is regarded as positive if its spatial IoU and temporal IoU with any ground truth are above $\tau$. Figure \ref{fig:recall} shows the performance of our tubelet generation model. The curves show that tracking based object linking achieves much better performance than greedy linking based method and achieves  87\% recall@IoU=0.3.

\section{Activity Localization and Classification}
We exploit Pseudo-P3D~\cite{qiu2017learning} framework as our backbone for tubelet action localization and classification. For input of P3D, the original version is the short clip of 16 consecutive frames. To capture the long-time actions, we chose to uniformly sample 64 frames across the temporal span of each proposal. The temporal span means sampling the same frames if the length of proposal is less than 64.
The target for P3D is to distinguish the label of proposals or the non-action class. A proposal with spatio Intersection of Union (IoU) with 0.35 and temporal IoU above 0.5 with the ground-truth is designated as a ``positive" proposal and assign an action class label.The proposals whose temporal IoU with ground-truth is less than 0.2 are regarded as ``negative" samples. In order to improve the robustness of our network, the online hard negative mining~\cite{ShrivastavaCVPR16} method is utilized during training.
It is reasonable to decompose the listed 18 classes into two equal parts, one for car-related activities and the other for person-related activities. Table \ref{table:1.1} details the action categories of each part. we train two of our P3D model for each group of activities and fuse localization results of each model.

\textbf{Experiment Results.} Table \ref{table:1.2} shows the performances of all the components in our trimmed action detection system. Overall, the tracking tubelets achieves the lower missing (84.13\%) value compared with the greedy linking. And by additionally applying the model decomposition to capture different actions in different categories, the whole model achieves an obvious improvement, which gets the lowest missing value (75.17\%) under the false alarm rate 0.15 in validation set. For the final submission, we also utilize the public post processing method soft-NMS~\cite{Bodla:ICCV17} to further improve the performance.

\section{Conclusion}
For ActEV-PC task in ActivityNet Challenge 2019, we mainly focused on action detection on spatial and temporal aspects with different tubelet generation and model decomposition methods. Our future works include more in-depth studies of how fusion weights of different components could be determined to boost the action detection and localization performance.

{
\bibliographystyle{ieeetr}
\bibliography{egbib}
}

\end{document}